\documentclass[letterpaper, 10 pt, conference]{ieeeconf}  % Comment this line out if you need a4paper
\IEEEoverridecommandlockouts                              % This command is only needed if 
\usepackage{graphics} % for pdf, bitmapped graphics files
\usepackage{amsmath} % assumes amsmath package installed
\usepackage{amssymb}  % assumes amsmath package installed
\usepackage{amsfonts}
\usepackage{xcolor}
\usepackage{epsfig}
\usepackage{graphicx}
\usepackage{subfigure}
\usepackage{epstopdf}
\usepackage{hyperref}
\usepackage{multirow}
\usepackage{booktabs}
\usepackage{array}
\usepackage{algorithm}
\usepackage{algorithmic}
\usepackage{textcase}
\usepackage{graphicx}
\usepackage{picinpar}
\usepackage{url}
\usepackage[latin1]{inputenc}
\usepackage{colortbl}
\usepackage{pifont}
\usepackage{color}
\usepackage{alltt}
\usepackage{enumerate}
\usepackage{siunitx}
\usepackage{breakurl}
\usepackage{pbox}
\usepackage{tikz}
\usepackage{caption}
\usepackage{threeparttable}
\usepackage{tabularx}
\usepackage{mathrsfs}
\usepackage{makecell}
\usepackage{pgfplots}
\usepackage{textcomp}
\usepackage{dsfont}
\usepackage{mathtools,xparse}
\usepackage[super]{nth}
\usepackage{soul}
\usepackage{cite}

\title{\LARGE \bf
A Large-Scale Dataset for Benchmarking Elevator Button Segmentation and Character Recognition}

\author{\authorblockN{Jianbang Liu$^{1}$, Yuqi Fang$^{1}$, Delong Zhu$^{1*}$, Nachuan Ma$^{1}$, Jin Pan$^{1}$, and Max Q.-H. Meng$^{2*}$}
\thanks{$^*$The corresponding author of this paper.} 
\thanks{$^1$The authors are with the Department of Electronic Engineering, The Chinese University of Hong Kong, Shatin, N.T., Hong Kong SAR, China. \textit{email: \{henryliu, fangyuqi, zhudelong, manachuan, jpan\}@link.cuhk.edu.hk}}
\thanks{$^2$Max Q.-H. Meng is with the Department of Electronic and Electrical Engineering, Southern University of Science and Technology, Shenzhen, China, and also with the Shenzhen Research Institute of the Chinese University of Hong Kong, Shenzhen, China, on leave from the Department of Electronic Engineering of the Chinese University of Hong Kong, Hong Kong \textit{e-mail: max.meng@ieee.org}. This project is partially supported by the Hong Kong RGC GRF grants \#14200618 and Hong Kong ITC ITSP Tier 2 grant \#ITS/105/18FP awarded to Max Q.-H. Meng.} 	
}

\begin{document}

\maketitle
\thispagestyle{plain}
\pagestyle{plain}
\pagenumbering{arabic}
\newcommand{\etal}{\textit{et al.}}
\definecolor{ao}{rgb}{0.0, 0.5, 0.0}
\makeatletter
% Reinsert missing \algbackskip
\def\algbackskip{\hskip-\ALG@thistlm}
\makeatother

\newcommand\mycommfont[1]{\footnotesize\ttfamily\textcolor{black}{#1}}
% \SetCommentSty{mycommfont}

% \SetNlSty{bfseries}{\color{black}}{}

%%%%%%%%%%%%%%%%%%%%%%%%%%%%%%%%%%%%%%%%%%%%%%%%%%%%%%%%%%%%%%%%%%%%%%%%%%%%%%%%
\begin{abstract}
Human activities are hugely restricted by COVID-19, recently. Robots that can conduct inter-floor navigation attract much public attention, since they can substitute human workers to conduct the service work. However, current robots either depend on human assistance or elevator retrofitting, and fully autonomous inter-floor navigation is still not available. As the very first step of inter-floor navigation, elevator button segmentation and recognition hold an important position. Therefore, we release the first large-scale publicly available elevator panel dataset in this work, containing 3,718 panel images with 35,100 button labels, to facilitate more powerful algorithms on autonomous elevator operation. Together with the dataset, a number of deep learning based implementations for button segmentation and recognition are also released to benchmark future methods in the community. The dataset is available at \url{https://github.com/zhudelong/elevator_button_recognition}
\end{abstract}

%%%%%%%%%%%%%%%%%%%%%%%%%%%%%%%%%%%%%%%%%%%%%%%%%%%%%%%%%%%%%%%%%%%%%%%%%%%%%%%%
\section{INTRODUCTION}
\label{sec_intro}
% \begin{enumerate}
%   \item \textcolor{red}{Emphasize the importance of button recognition in the autonomous elevator operation. }
%   \item \textcolor{red}{Briefly introduce the dataset (object detection and semantic segmentation are boosted by the deep learning technology. Deep learning requires large scale data to achieve human level performance...) and experiment conducted (ablation study of semantic segmentation algorithm).}
%   \item \textcolor{red}{Introduce the designed button segmentation and recognition framework.}
%   \item \textcolor{red}{Introduce the structure of the paper content.}
% \end{enumerate}
Affected by the coronavirus, COVID-19, human activities are hugely restricted, e.g., people in those severe epidemic areas are forced to stay at home. The autonomous robot systems draw more public attention because they can substitute human workers and conduct many service work. Take the hospital as an example, if a mobile robot can move freely and safely across floors, it can replace the healthcare workers to deliver drugs to the infectious patients, greatly reducing the risk and burden on the medical personnel. However, fully autonomous inter-floor navigation of the service robot is still quite challenging nowadays.

\begin{figure}[t]
	\centering
	\includegraphics[width=0.4\textwidth]{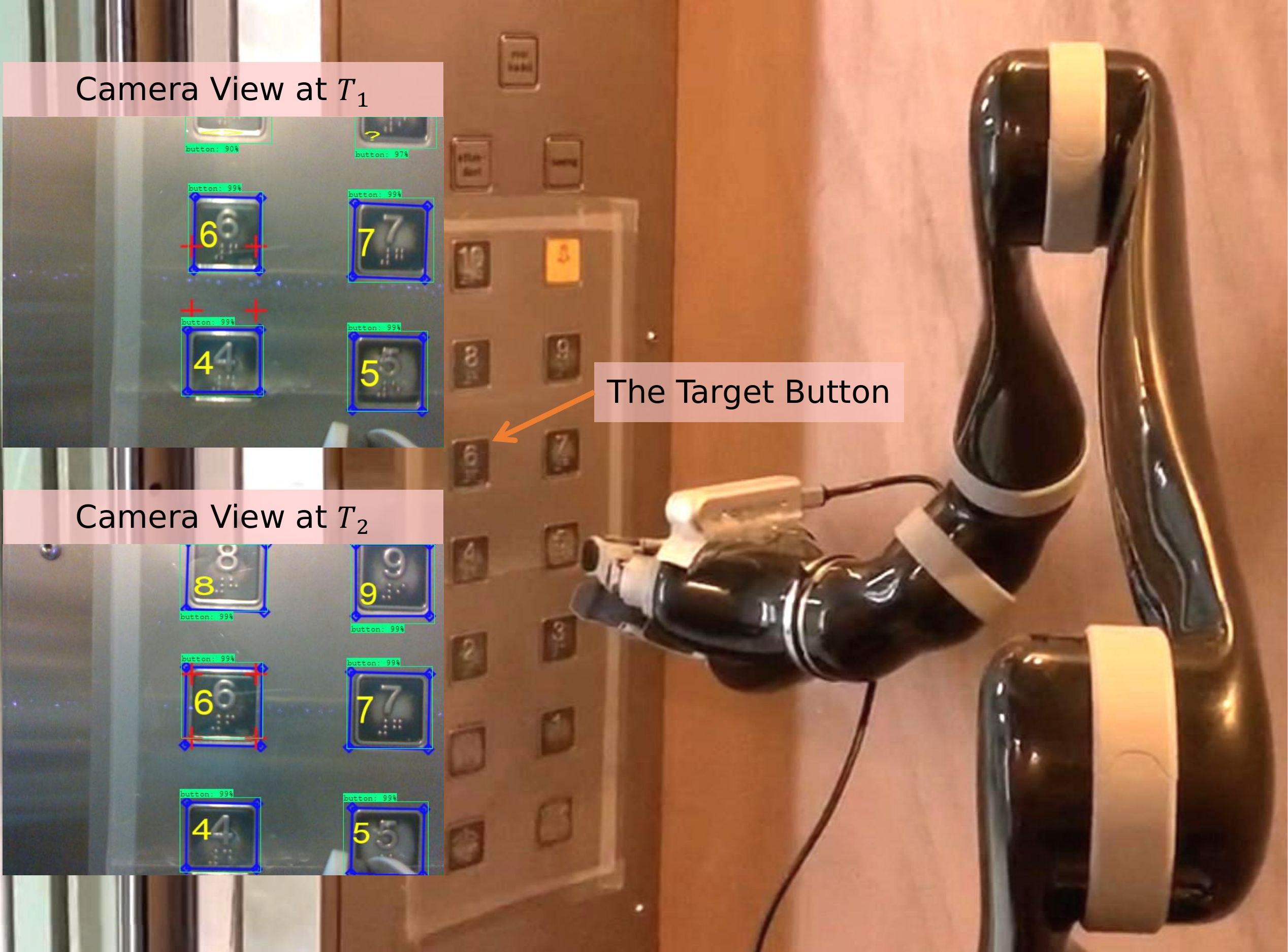}
	\caption{Demonstration of elevator operation. The robot arm is tuning the pose of its end-effector and trying to press the target button.}
	\label{fig_robot_operate_elevator}
	\vspace{-2mm}
\end{figure}

To navigate the robot to travel across floors, traditional approaches either rely on assistance from the outside world or are simply based on the hand-craft features under specific environments \cite{miura2005interactive,kang2007navigation,klingbeil2010autonomous,kim2011robust}, which make them not generally applicable. As the very first and fundamental step of autonomous inter-floor navigation, elevator button segmentation and recognition play an important role, which can enormously affect the performance of the robot system (see Fig.~\ref{fig_robot_operate_elevator}). However, there exists no large-scale elevator panel dataset for the researchers to verify their button segmentation and recognition approaches. As shown in Table~\ref{table:dataset}, existing elevator datasets \cite{klingbeil2010autonomous, zakaria2014elevator, liu2017recognizing, yang2018intelligent} either contain a few panels images or do not make data available. Therefore, we in this work release a large-scale elevator panel dataset to facilitate related studies on autonomous elevator operation. Moreover, benefiting from deep learning techniques that automatically mine informative features, a number of learning-based network implementations on button segmentation and recognition are also released to benchmark future methods in the community.

The contributions of this paper are summarized as follows:
\begin{itemize}
  \item The first large-scale publicly available elevator panel dataset is released in this work, containing high-quality 3,718 panel images with 35,100 button labels.
  \item The baseline implementations on button segmentation and recognition, as well as evaluation metrics, are also released, for benchmarking future methods and facilitating related studies on autonomous elevator operation.
\end{itemize}

The remainder of this paper is organized as follows. Section~\ref{related_work} briefly surveys existing elevator panel datasets and recent studies on button operation. The details and characteristics of the released dataset are described in Section \ref{sec_dataset}. The evaluation metrics are established in Section~\ref{sec_metrics}, and network implementations for button segmentation and recognition in Section~\ref{sec_experiment}. The paper is concluded in Section~\ref{sec_conclusion}.

\section{RELATED WORK}
\label{related_work}
\subsection{Elevator Panel Dataset}
\begin{table*}[t]
	\caption{The Comparison between our released elevator panel dataset and existing panel datasets. ``-'' denotes \textit{not reported}. }
	\label{table:dataset}
	\centering
	\begin{tabular}{lc|c|c|c|cccl}
		\toprule
		\multirow{2}{*}{} & \multicolumn{4}{c}{Dataset Statistic} & \multicolumn{4}{c}{Dataset Feature} \\ \cmidrule(l){2-5} \cmidrule(l){6-9} 
		Dataset & \multicolumn{1}{r|}{Images} & \multicolumn{1}{l|}{Buttons} & \multicolumn{1}{l|}{panels} & \multicolumn{1}{l|}{Classes} & \multicolumn{1}{l|}{Video Sequence Image} & \multicolumn{1}{l|}{GT Pose} & \multicolumn{1}{l|}{Segmentation Map}  & public  \\ \midrule
		Klingbeil \textit{et al.} \cite{klingbeil2010autonomous} & 150 & 686 & 60 & - & \multicolumn{1}{c|}{$\times$}  & \multicolumn{1}{c|}{$\times$} & \multicolumn{1}{c|}{$\times$} & $\times$ \\
		Zakaria \textit{et al.} \cite{zakaria2014elevator} & 50  & - & 15+ & - & \multicolumn{1}{c|}{$\times$}  & \multicolumn{1}{c|}{$\times$} & \multicolumn{1}{c|}{$\times$} & $\times$ \\
		Liu and Tian \cite{liu2017recognizing}  & 1,000 & - & - & - & \multicolumn{1}{c|}{$\times$}  & \multicolumn{1}{c|}{$\times$} & \multicolumn{1}{c|}{\checkmark} & $\times$ \\
		Yang \textit{et al.} \cite{yang2018intelligent} & 260,560 & -  & 8 & 26 & \multicolumn{1}{c|}{$\times$}  & \multicolumn{1}{c|}{$\times$} & \multicolumn{1}{c|}{$\times$} & $\times$ \\
		Ours & 3,718 & 35,100 & 2,100+ & 297 & \multicolumn{1}{c|}{\checkmark}  & \multicolumn{1}{c|}{\checkmark} & \multicolumn{1}{c|}{\checkmark} & \checkmark \\ \bottomrule
	\end{tabular}
	\vspace*{-3mm}
\end{table*}
\begin{table*}[t]
	\centering
	\newcolumntype{M}[1]{>{\centering\arraybackslash}m{#1}}
	\begin{tabular}{lM{27mm}M{27mm}M{27mm}M{27mm}M{27mm}}
		\toprule
		& Sample 1                                                  & Sample 2                                                  & Sample 3                                                  & Sample 4                                                           & Sample 5                                    \\ 
		\midrule
		Original        & \includegraphics[width=9em]{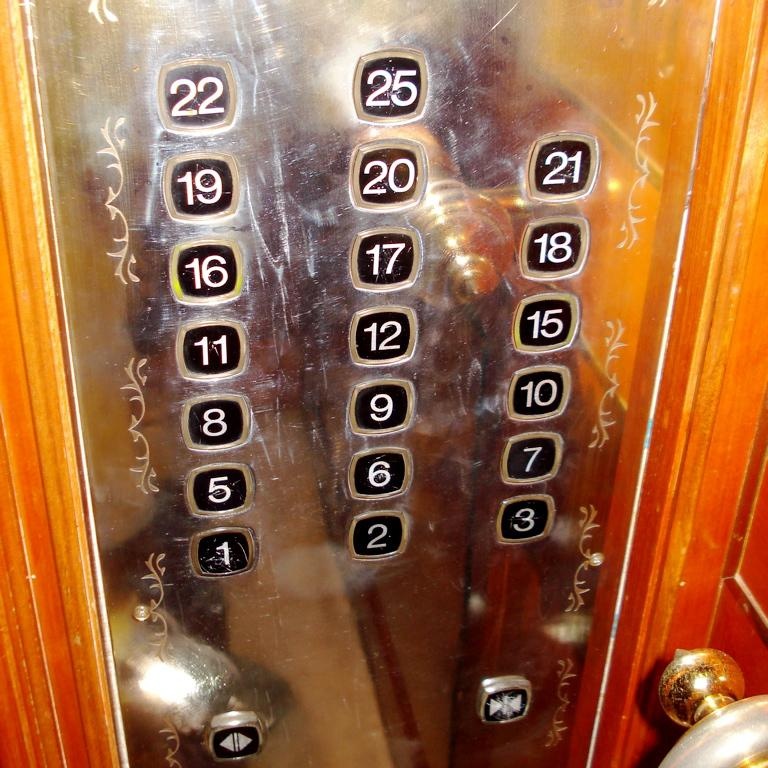} & \includegraphics[width=9em]{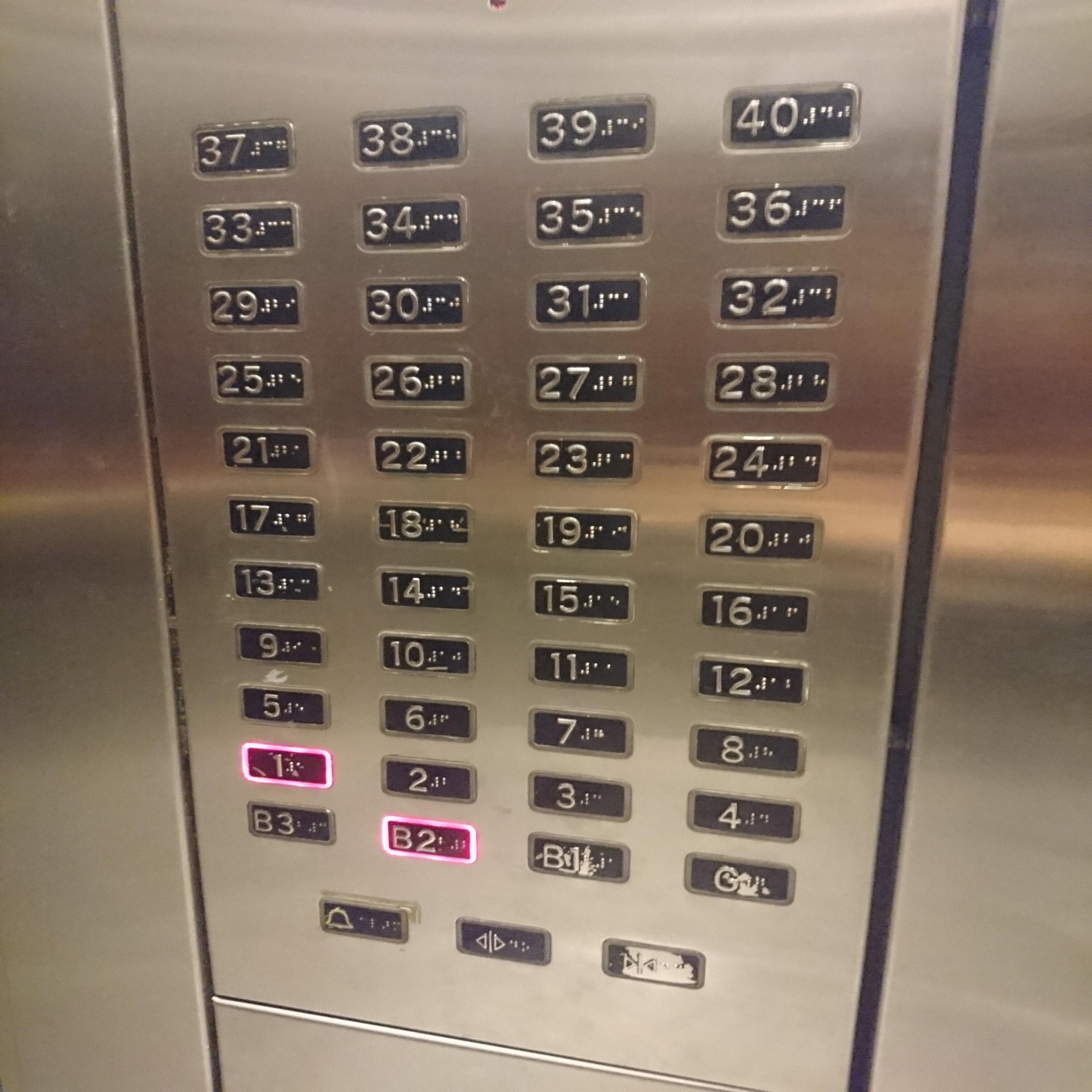} & \includegraphics[width=9em]{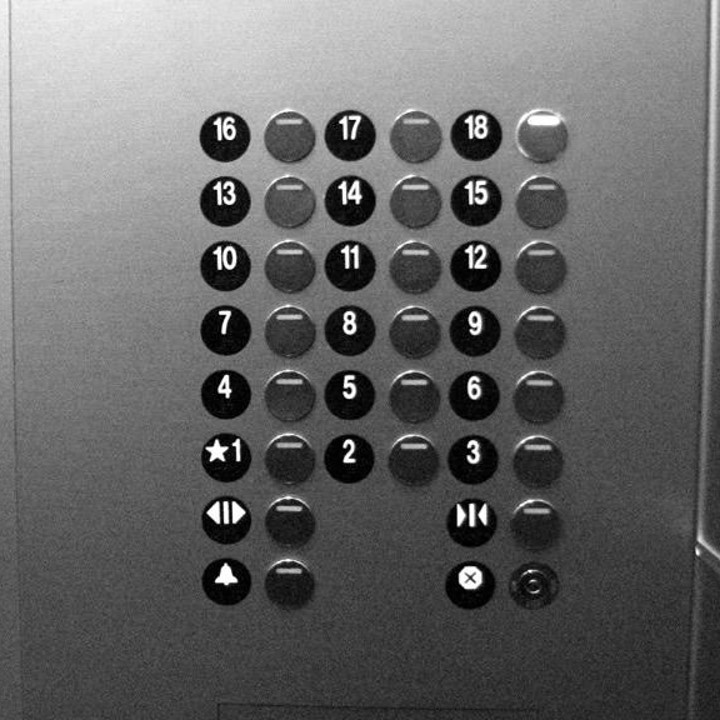} & \includegraphics[width=9em]{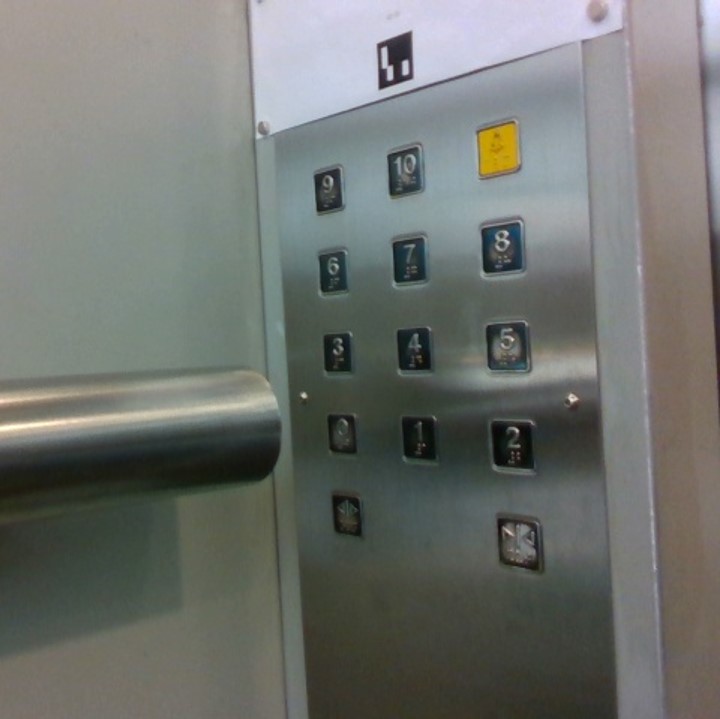} & \includegraphics[width=9em]{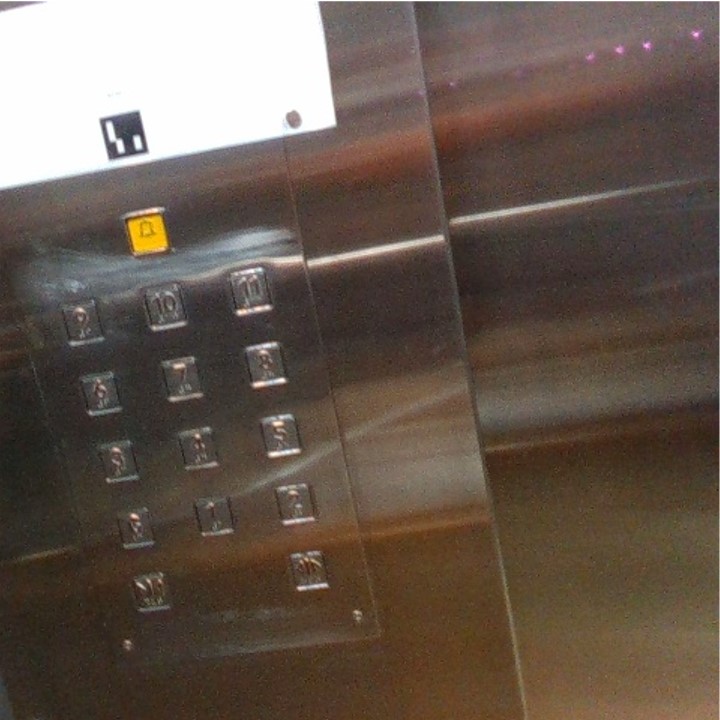} \\
		Segmentation GT & \includegraphics[width=9em]{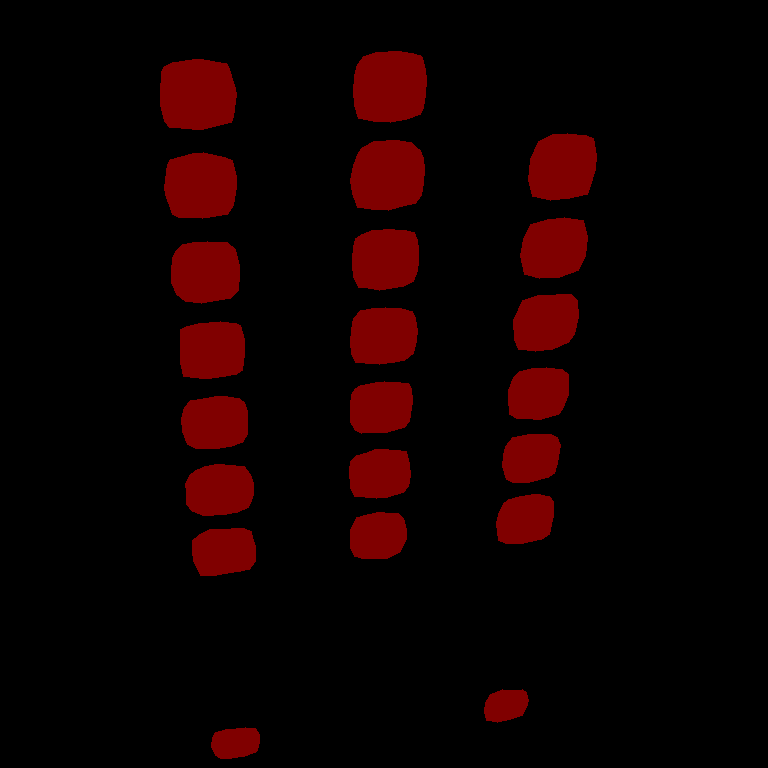} & \includegraphics[width=9em]{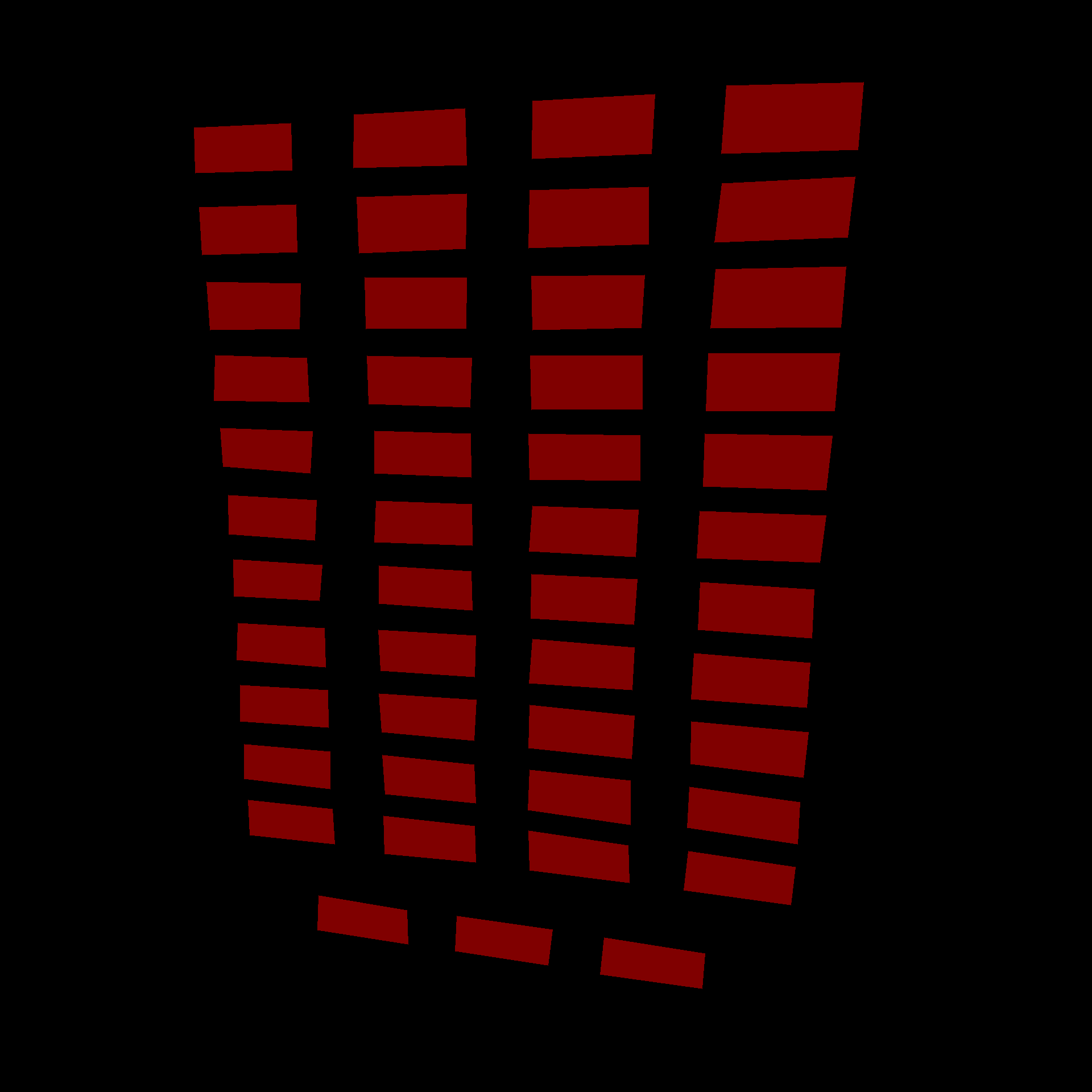} & \includegraphics[width=9em]{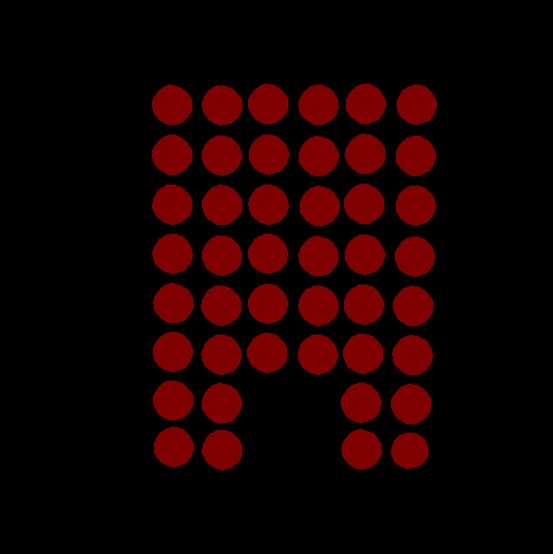} & \includegraphics[width=9em]{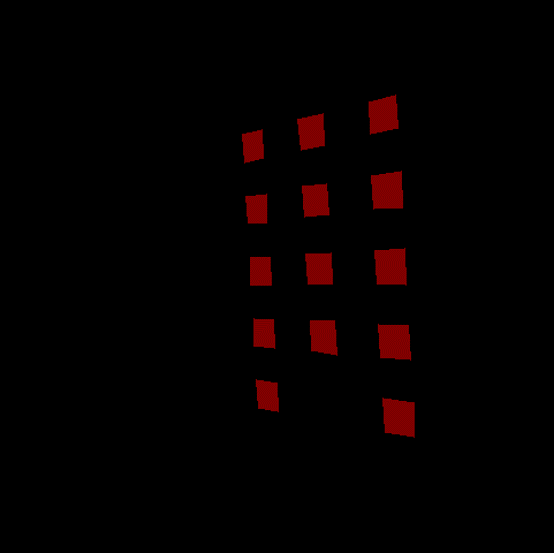} & \includegraphics[width=9em]{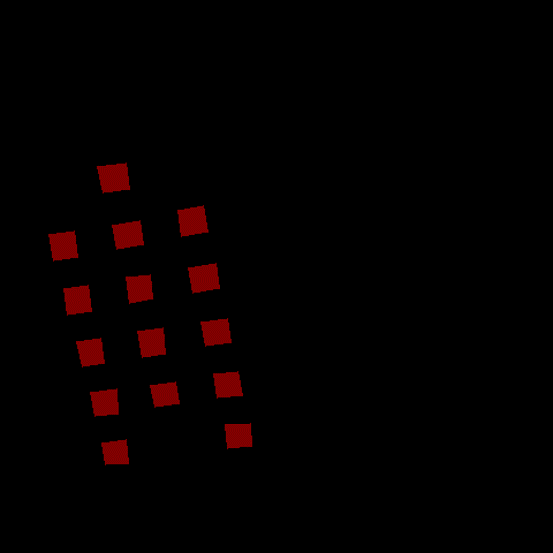} \\
		Recognition GT    & \includegraphics[width=9em]{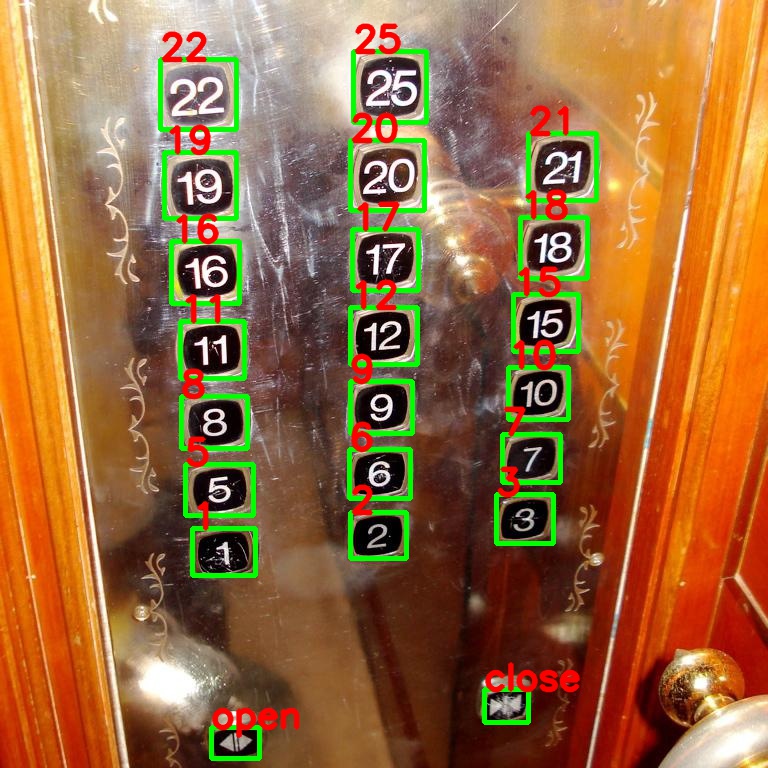} & \includegraphics[width=9em]{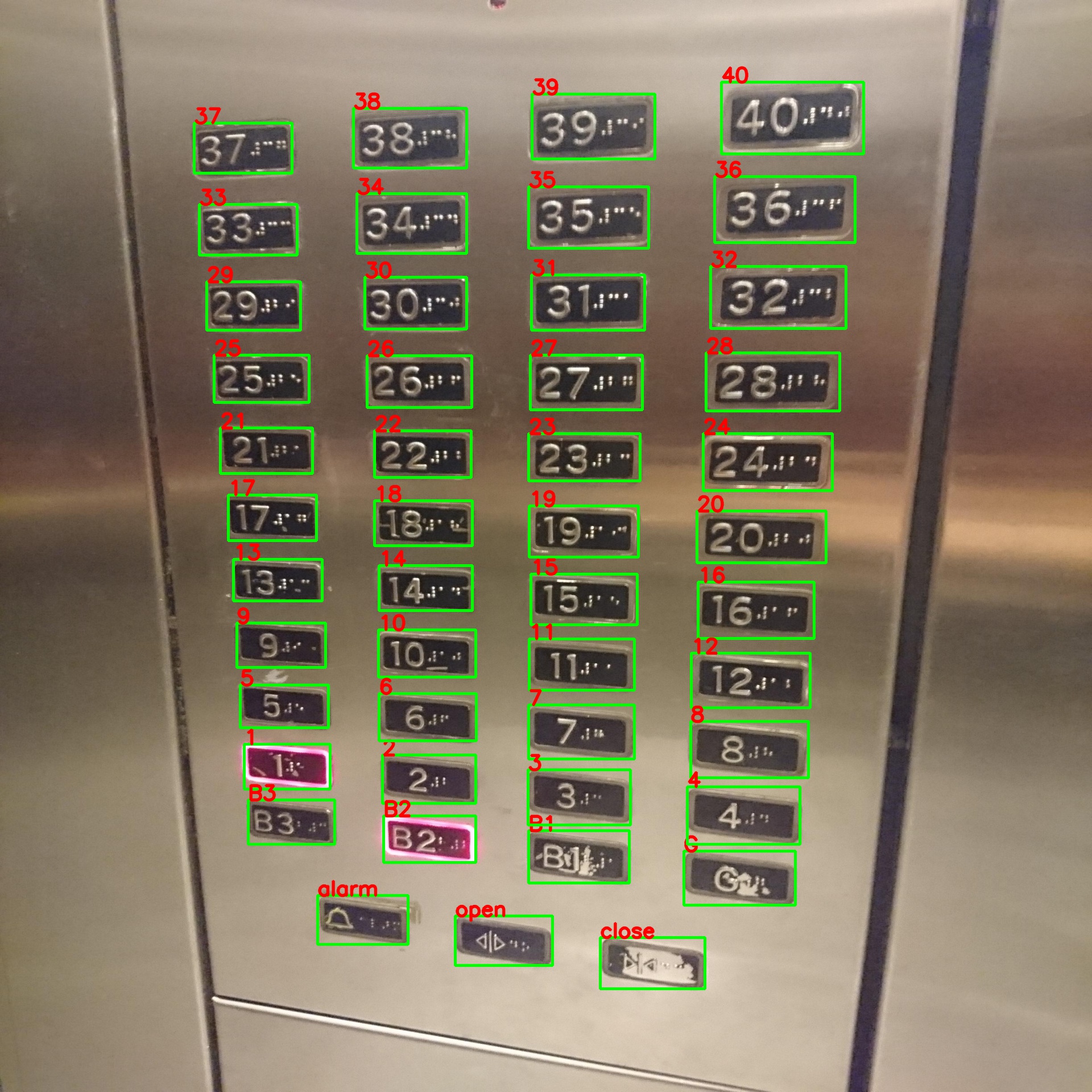} & \includegraphics[width=9em]{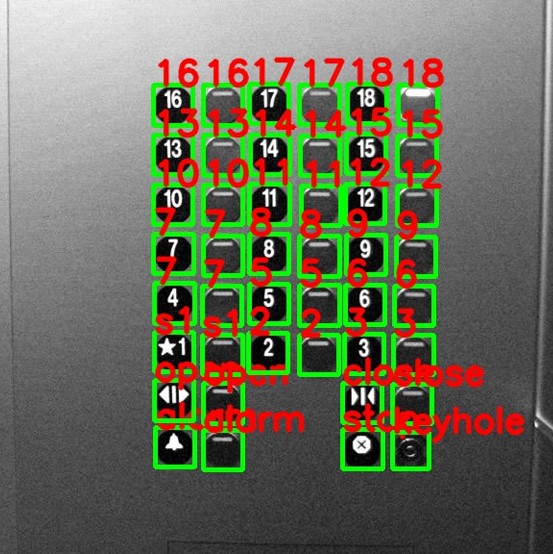} & \includegraphics[width=9em]{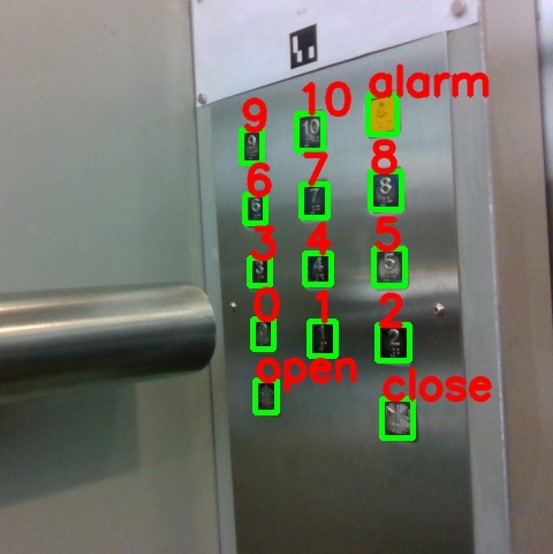} & \includegraphics[width=9em]{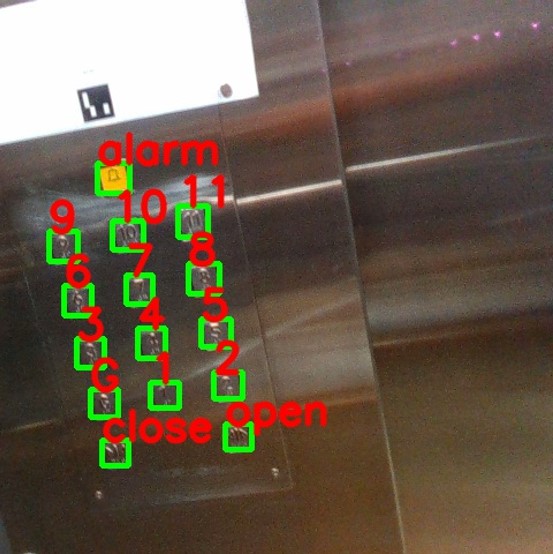} \\
		\bottomrule
	\end{tabular}
	\captionof{figure}{Examples of the elevator panel samples with pixel-wise segmentation masks (row 2) and button recognition labels (row 3).}
	\label{fig_stat_imgs}
	\vspace{-4mm}
\end{table*}

In the field of computer vision, some large-scale image datasets have been released to encourage the researchers to develop deep learning based solutions, such as ImageNet \cite{ILSVRC15}, KITTI \cite{Menze2015CVPR}, and PASCAL VOC \cite{Everingham10}. As we know, with more sufficient training data, the generalization ability and robustness of a deep learning neural network can be improved. However, in autonomous elevator operation scenarios, there exists no public large-scale elevator panel dataset for button segmentation and recognition. For instance, Klingbeil \textit{et al.} \cite{klingbeil2010autonomous} collect only 150 panel images from more than 60 distinct elevators for button localization and optical character recognition (OCR). The training and testing sets are only 100 and 50 panel images, respectively. Zakaria \textit{et al.} \cite{zakaria2014elevator} evaluate their proposed button recognition and detection system using 50 high-resolution images including the internal and external elevator panels. Liu \textit{et al.} \cite{liu2017recognizing} collect 1,000 panel images captured from both inside and outside of the elevators, to train the deep learning based method for pixel-wise button segmentation. The biggest panel dataset to date is collected by Yang \textit{et al.} \cite{yang2018intelligent}, which contains 260,560 images that are composed of 8 different panels, and many data augmentation strategies have been utilized to further enlarge the data scale, e.g., blurring, sharpening, and histogram equalization. However, Yang \textit{et al.} \cite{yang2018intelligent} do not release their collected dataset, thus no comparison studies can be conducted. This paper releases the first public large-scale elevator panel dataset, which contains 3,718 panel images and 35,100 button labels, covering the majority of existing button categories. Furthermore, our dataset contains both static images and video sequences, as well as pose information between the panels and the vision camera. We believe that this dataset can benchmark future methods on autonomous elevator operation.

\subsection{Autonomous Elevator Operation}

\begin{table*}[t]
	\centering
	\newcolumntype{M}[1]{>{\centering\arraybackslash}m{#1}}
	\begin{tabular}{lM{140mm}}
		\toprule
		& \includegraphics[width=100mm]{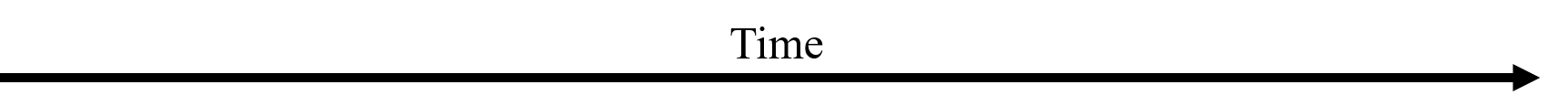}   \\
		\midrule
		Original        & \includegraphics[width=130mm]{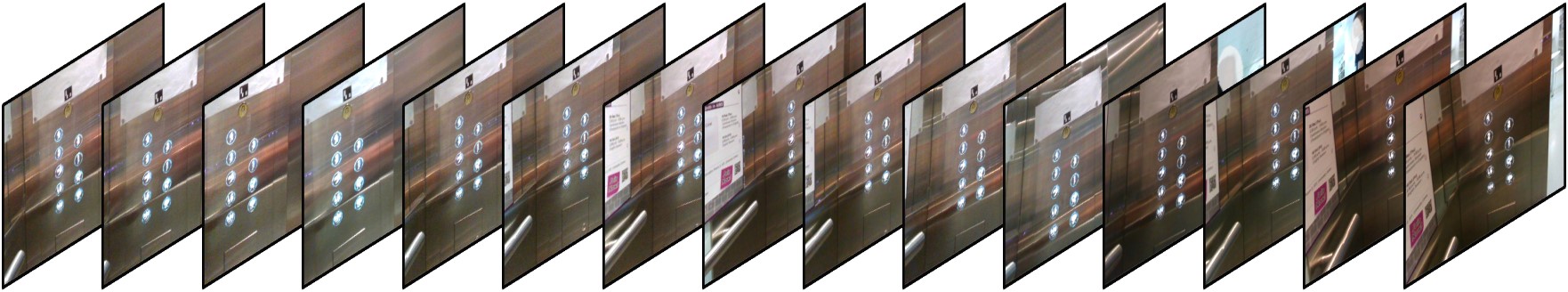}    \\
		Segmentation GT & \includegraphics[width=130mm]{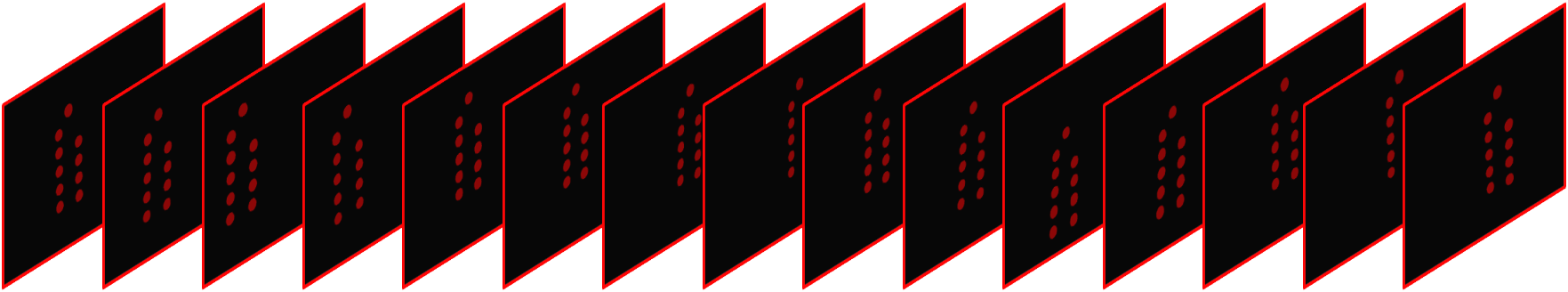}   \\
		Recognition GT    & \includegraphics[width=130mm]{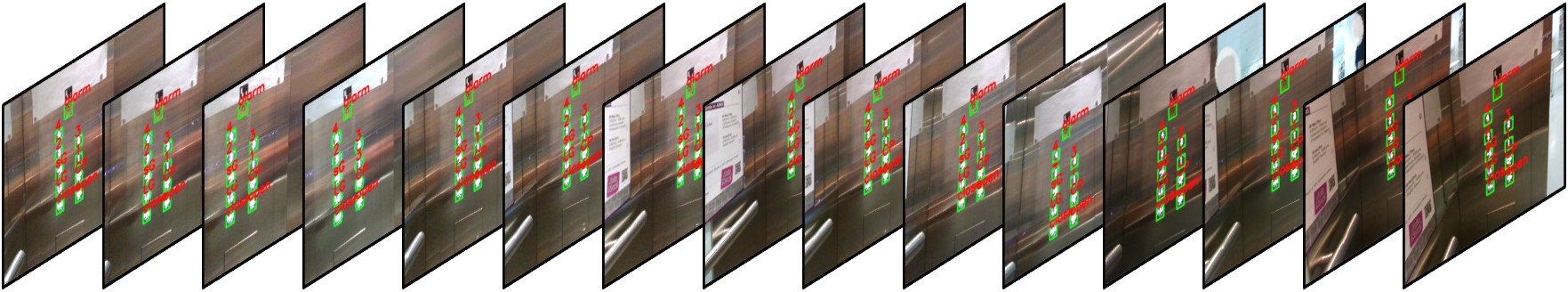} \\
		\bottomrule
	\end{tabular}
	\captionof{figure}{The video sequence samples with ground-truth segmentation masks and button recognition labels in our dataset.}
	\label{fig_video_seq_imgs} 
	\vspace{-2mm}
\end{table*}

Research on elevator button segmentation and recognition has received more attention in recent years. The existing methods can be categorized into traditional methods and deep learning based methods. For traditional methods, as one of the earliest studies on autonomous elevator operation, Klingbeil \textit{et al.} \cite{klingbeil2010autonomous} perform button detection using a sliding-window detector first, and then apply the Expectation Maximization (EM) algorithm to remove false positives and infer missing buttons. In button recognition stage, the OCR model is adopted, and then a hidden markov model (HMM) is leveraged to take into consideration the arrangement of button labels in order of floors. Their method correctly detects and labels $86.2\%$ of the buttons in a small-scale dataset with 50 test panels. Kim \textit{et al.} \cite{kim2011robust} propose a robust vision-based button recognition method, which is comprised of feature extraction, initial button recognition, and post-processing modules. Moreover, considering that a button object has a convex quadrilateral boundary, the authors design the specific features to represent the button contour under different perspective distortions, which overcomes the difficulty caused by the reflective walls or the partial occlusion and greatly improves button segmentation accuracy. Zakaria \textit{et al.} \cite{zakaria2014elevator} develop an efficient button detection and recognition framework based on the sobel operator \cite{kanopoulos1988design} for button edge detection and Wiener filter for reflection noise removal. These traditional algorithms highly rely on the domain knowledge and manual design, and most of them are only evaluated in some specific environments, hence their performance and robustness remain unsatisfying.

Benefiting from deep learning technology, convolutional neural networks (CNNs) and recurrent neural networks (RNNs) have shown superior performance over traditional methods in different applications. Researchers have applied deep learning techniques to solve the button detection and recognition problem. Liu \textit{et al.} \cite{liu2017recognizing} adopt the fully convolutional network (FCN) for pixel-level segmentation of elevator buttons, and integrate the single-shot detector (SSD) with the convolutional recurrent neural network for button localization and recognition. Dong \textit{et al.} \cite{dong2017autonomous} first propose the proper button candidates based on R-CNN \cite{girshick2014rich}, and then a fine-tuned CNN is developed for elevator button recognition, which achieves a reliable and promising recognition performance. Based on the system proposed in \cite{dong2017autonomous}, Zhu \textit{et al.} \cite{zhu2018novel, 9324975} further modify the system by combining OCR network and Faster R-CNN network \cite{ren2015faster} into a single architecture, and thus button detection and recognition can be performed simultaneously, enabling an end-to-end training scheme. Yang \textit{et al.} \cite{yang2018intelligent} utilize the YOLO v2 network \cite{redmon2016yolo9000} for elevator button detection, and this 2D YOLO detector can produce the 3D coordinates of the target button based on a coordinate transform neural network. In this work, along with the released dataset, we introduce many classic and popular segmentation and recognition approaches, benchmarking and facilitating related studies on autonomous elevator operation.

\begin{figure}[b]
	\vspace{-4mm}
	\centering
	\includegraphics[width=0.46\textwidth]{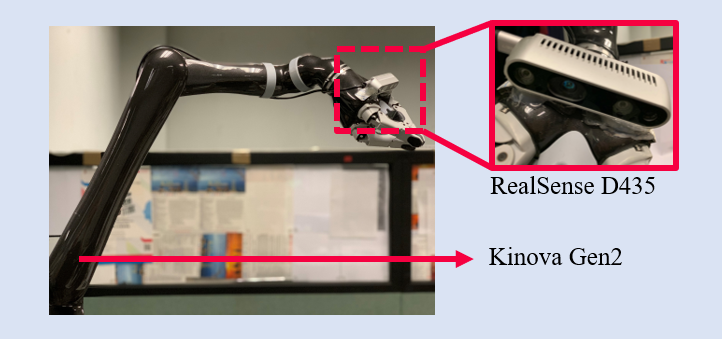}
	\caption{The hardware system for video sequence acquisition.}
	\label{fig_system_pic}
\end{figure}

\section{ELEVATOR BUTTON DATASET}\label{sec_dataset}

The released elevator panel dataset contains 3,718 panel images with 35,100 button labels. The images are comprised of two types: static images (Fig.~\ref{fig_stat_imgs}) and video sequences (Fig.~\ref{fig_video_seq_imgs}). The static images, which contains inner and outer control panels, are mostly collected from the Internet using a web crawler, and part of them are captured by the authors. We also manually clean the images to ensure there are no repetitive samples for the same elevator panel. The collected images are diverse in panel types, shooting angles, and lighting conditions.
The video sequences are captured by a RealSense depth camera D435 mounted on the end effector of a Kinova robot arm (see Fig.~\ref{fig_system_pic}). Specifically, we control the robot arm to move smoothly along some predefined trajectories, e.g., moving towards or away from the target panels while keeping them within the perception field. Images in the video sequences show slight blurs due to the movement of the robot arm, but this can increase the diversity of the training set, improving the generalization ability and robustness of the models.
% can simulate the image data streamed during the robot arm is searching for the button panel.
     
Furthermore, during image sequence acquisition, an ArUco marker \cite{garrido2014automatic} is sticked on each elevator panel, based on which the poses between the camera and the elevator panel can be performed with the help of OpenCV libraries. In \cite{zhu2019autonomous}, we utilize the pose information to assess the accuracy of the proposed distortion removal algorithm. The pose information is also released along with the elevator panel dataset, but in this work, it has not been exploited to assist button segmentation and recognition. To the best of our knowledge, this is the first dataset that integrates the pose information of elevator panels. We encourage future studies to leverage this prior knowledge to further improve the performance in autonomous button segmentation and recognition. The pose information also provides ground truth for single-image based pose estimation algorithms, which is the key for manipulation control. 

\subsection{Button Segmentation}
The collected elevator buttons have various shapes and appearances, e.g., circle, ellipse, rectangle, and trapezium, under different camera views. All buttons are carefully annotated. Specifically, we first draw a contour exactly outlining each button, and then generate the mask accordingly. All the mask labels are checked by colleagues in this research field and ascertain the correctness. 
In Fig \ref{fig_stat_imgs}, the segmentation ground truth is represented by binary masks, i.e., the dark red areas. Since the button class is annotated simultaneously, multi-class segmentation techniques can also be leveraged to address the problem.

\subsection{Button Recognition}
\begin{figure}[t]
	\centering
	\includegraphics[width=0.42\textwidth]{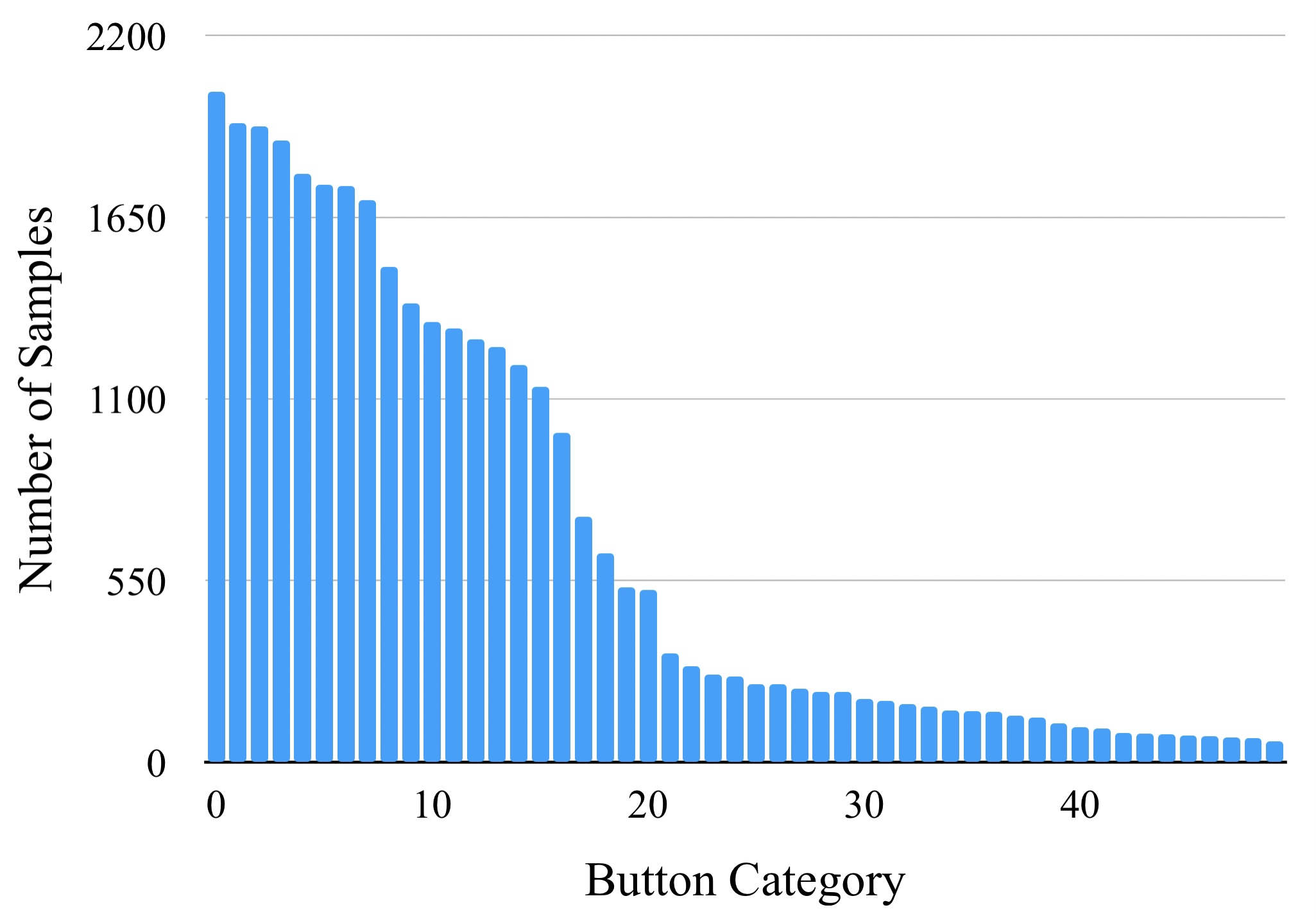}
	\caption{The class distribution of top 50 button categories.}
	\label{fig_distri_class}
	\vspace{-6mm}
\end{figure}

The released dataset contains 297 classes of buttons, covering the majority of existing button categories, which are labeled carefully by the authors. The distribution of button samples in top-fifty classes is demonstrated in Fig.~\ref{fig_distri_class}, indicating that there exists a severe class imbalance problem in the dataset. For example, some buttons (e.g., open, close, alarm), are commonly seen on most elevator panels, the number of which is unsurprisingly larger than buttons that are specifically designed, e.g., podium floor, rooftop, and ballrooms. Moreover, low-rise buildings are always more than high-rise ones, hence buttons with small numbers (e.g., floor 1, 2, 3) are significantly more than those with large numbers (e.g., floor 100). 

The class imbalance problem makes button recognition models easily overfitted to classes with sufficient training samples. 
To tackle this problem, we formulate the recognition task as an OCR problem instead of a classification problem. An OCR framework is developed in \cite{9324975}, which can decompose the text on a button into a series of characters.
For instance, floor $102$ can be split into $(1, 0, 2)$, and each character can be recognized individually. A character dictionary is specially designed for this dataset (see Table~\ref{table:char_dict}), based on which the OCR framework can recognize hundreds of button categories without changing the network design. 
%The details of the proposed OCR recognition scheme are described in \ref{sec_experiment}.

\begin{table}[t]
	\renewcommand{\arraystretch}{1.2}
	\newcolumntype{N}{>{\centering\arraybackslash}m{\dimexpr.135\linewidth-0.8\tabcolsep}}
	\caption{The designed character dictionary for the dataset. The functional buttons are encoded by the special characters, e.g., $><$ for \emph{close}, \$ for \emph{alarm}, \# for \emph{stop}, and \& for \emph{call}.}
	\label{table:char_dict}
	\centering
	\begin{tabular}{N N N N N N}
		\hline\hline \\[-3mm]
		`0': 0  & `1': 1   & `2': 2  & `3': 3  & `4': 4   & `5': 5 \\
		`6': 6  & `7': 7   & `8': 8  & `9': 9  & `A': 10  & `B': 11 \\
		`C': 12 & `D': 13 & `E': 14 & `F': 15 & `G': 16  & `H': 17 \\
		`I': 18 & `J': 19 & `K': 20 & `L': 21 & `M': 22  & `N': 23 \\
		`O': 24 & `P': 25 & `R': 26 & `S': 27 & `T': 28  & `U': 29 \\
		`V': 30 & `X': 31 & `Z': 32 & `$<$': 33 & `$>$': 34  & `(': 35 \\
		`)': 36 & `\$': 37 & `\#': 38 & `\&': 39 & `s': 40  & `-': 41 \\
		`*': 42 & `\%': 43 & `?': 44 & `!': 45 & `$\phi$': 46 \\
		[0.6ex]
		\hline\hline
	\end{tabular}
\end{table} 

%Take what we have done in \cite{zhu2018novel} as a reference, the multi-branch model has classification branch and an OCR branch at the final stage. The category of buttons can be treated as the ground truth for classification branch and the exact classes of buttons will be the ground truth for the OCR branch's prediction.          
          
% We collect and annotate the images of internal and external elevator control panels. In advance, we construct this large-scale dataset by merging our collected data with others public button panel image datasets, which the format of the label information has been revised for data consistency.
% The label information used to describe every button contains the class name of that button belonging to and the image coordinates of some control points, which lay on the boundary of the button and form the shape of the button by connecting them in sequence.

\begin{figure*}[t]
	\centering
	\includegraphics[width=\textwidth]{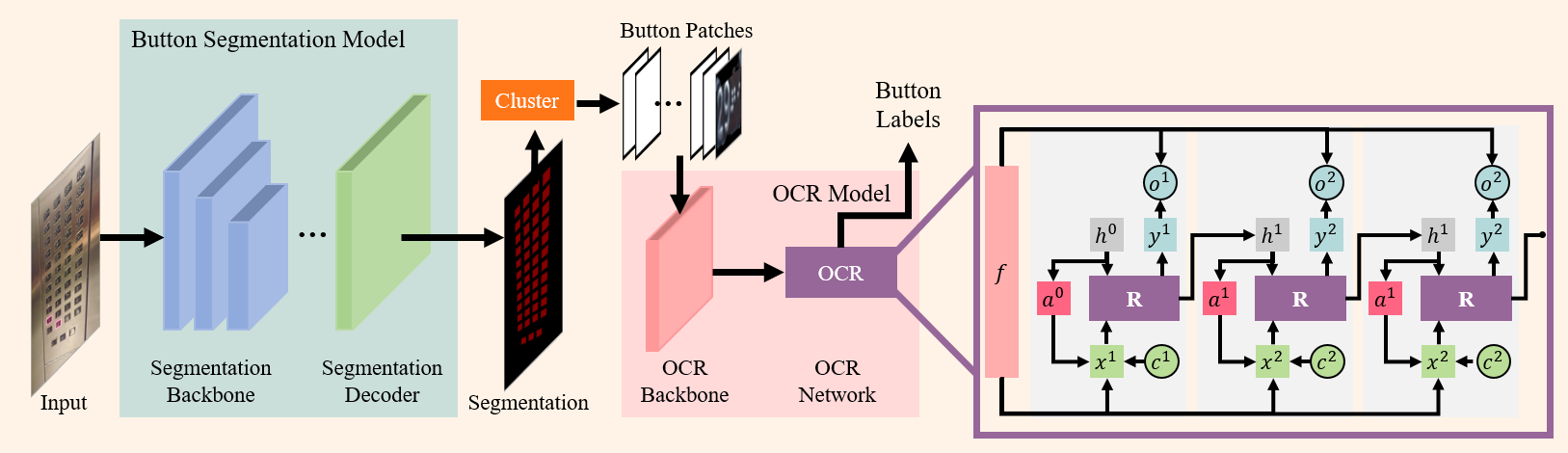}
	\caption{The proposed button segmentation and recognition framework. The segmentation model takes a RGB panel image as the input, and generate the button segmentation mask. A clustering module (DBSCAN) aims to group pixels close to each other as an independent button, and frames the button with a tight bounding box. Based on each box, the OCR model implemented via an attention RNN network (right) recognizes its button character in sequence. The details of the OCR model can refer to \cite{zhu2018novel}.}
	\label{fig_framework}
	\vspace{-2mm}
\end{figure*}

\section{BENCHMARK METRICS}\label{sec_metrics}
% \begin{enumerate}
%   \item \textcolor{red}{Describe the evaluation metrics of segmentation.}
%   \item \textcolor{red}{Describe the evaluation metrics of character recognition.}
% \end{enumerate}

In this section, several commonly used metrics are introduced to evaluate the performance of elevator button segmentation and recognition among different algorithms.

\subsection{Metrics for Button Segmentation}
Four measurement metrics are adopted to validate the button segmentation performance, i.e., \textit{Intersection over Union}, \textit{Precision}, \textit{Recall}, and \textit{Parameter Size}, each of which is detailed as follows.

\textbf{Intersection over Union}, also known as Jaccard Index, is the most popular metric for evaluation of the segmentation performance, which aims to measure the overlap between the predicted segmentation masks and the ground-truth areas. For a given input panel image, IoU is calculated as follows:

\begin{equation}
IoU = \frac{TP}{TP + FP + FN},
%IoU = \frac{f_{i} \cap g_{i}}{f_{i} \cup g_{i}},
\end{equation}
%where $f_{i}$ indicates the probability of pixel $i$ being categorized as button, and $g_{i}\in$\{1, 0\} means the corresponding label of the ground-truth area.
where True Positive (TP) indicates the number of pixels that are correctly classified as button pixels; False Positive (FP) means the number of pixels that are wrongly classified as button pixels; False Negative (FN) denotes the number of pixels that are wrongly classified as the background. 

\textbf{Precision and Recall} are also two common metrics for segmentation quality measurement. Precision is sensitive to the over-segmentation, while recall is sensitive to the under-segmentation, which are denoted as follows:
\begin{equation}
Precision = \frac{TP}{TP + FP}, \; \; Recall = \frac{TP}{TP + FN}.
\end{equation}

\textbf{Parameter Size}, measuring the number of training parameters, is an extremely important metric in button segmentation task. Although computation memory is usually sufficient, it can be a limiting factor in some scenarios, especially those requiring the real-time manipulation. In these situations, models with smaller parameter size can be extraordinarily helpful, enabling more efficient elevator operation. However, it is possible that these light-weighted models produce less satisfactory performance compared with the heavy ones, hence a trade-off between the efficiency and accuracy needs to be made in practice.

\subsection{Metrics for Button Recognition}
For button recognition, $Average\ Accuracy\ (ACC)$ is adopted to evaluate the overall recognition performance. $ACC$ is the percentage of the correctly recognized buttons out of the total amount of buttons involved, which is:
\begin{equation}
ACC = \frac{No.\ of\ button\ recognized\ correctly}{No.\ of\ button\ involved}.
\end{equation}

\begin{table}[b]
	\centering
	\vspace{-4mm}
	\caption{The button segmentation results based on various algorithms. Abbreviations: IoU: intersection over union; Params: parameter size; Res50, Mob2, Mob3, and Gh represent ResNet50, MobileNetV2, MobileNetV3, and GhostNet, respectively.}
	\label{table:seg}
	\begin{tabular}{@{}lcccc@{}}
		\toprule
		& IoU(\%)         & Precision(\%)  & Recall(\%)      & Params(M)     \\ \midrule
		DeeplabV3+(Res50) & 53.40          & 55.16          & 94.36           & 40.35         \\
		DeeplabV3+(Mob2)  & 34.78          & 35.68          & 93.25           & 5.82          \\
		DeeplabV3+(Mob3)  & 47.30          & 48.84          & 93.76           & 9.64          \\
		DeeplabV3+(Gh)    & 53.34          & 56.25          & 91.16           & 8.25          \\
		FCN32s            & 51.57          & 64.24          & 72.33           & 134.27        \\
		FCN16s            & 62.45          & \textbf{74.01} & 80.00           & 134.27        \\
		FCN8s             & \textbf{62.72} & 73.75          & 80.74           & 134.27        \\
		ICNet             & 46.93          & 55.94          & 74.44           & 7.73          \\
		ENet              & 42.36          & 43.25          & \textbf{95.39}  & \textbf{0.35} \\
		U-Net             & 57.65          & 67.36          & 80.00           & 1.94          \\
		PSPNet            & 56.54          & 60.10          & 90.54           & 68.06         \\ 
		\bottomrule
	\end{tabular}
	\vspace{-2mm}
\end{table}

\begin{table*}[t]
	\caption{The button recognition results on the test set (558 images and 5,309 buttons) based on different segmentation algorithms described in Section~\ref{sec_exp_seg}. ``Button Detected": the number of buttons correctly detected;	``ACC (Detected)" and ``ACC (All)" represent the percentage of correctly recognized buttons out of the detected buttons, and the whole test buttons, respectively.}
	\label{table:recog}
	\centering
	\begin{tabular}{@{}lcccccccc@{}}
		\toprule
		
		\multirow{2}{*}{}     & \multicolumn{3}{c}{Additive OCR Model}                                               & & \multicolumn{3}{c}{Multiplicative OCR Model}                              \\ \cmidrule(lr){2-4} \cmidrule(l){6-8} 
		Segmentation Model    & \multicolumn{1}{c}{Button Detected} & ACC (Detected)    & \multicolumn{1}{c}{ACC (All)} & & \multicolumn{1}{c}{Button Detected} & ACC (Detected)    & ACC (All)         \\ \midrule
		DeeplabV3+ (Res50)    & 3,296                               & \textbf{73.74\%} & 62.08\%                      & & 3,251                               & \textbf{72.73\%} & 61.24\%          \\
		DeeplabV3+ (Mob2)     & 1,983                               & 69.58\%          & 37.35\%                      & & 1,957                               & 68.67\%          & 36.86\%          \\
		DeeplabV3+ (Mob3)     & 2,781                               & 72.69\%          & 52.38\%                      & & 2,705                               & 70.70\%          & 50.95\%          \\
		DeeplabV3+ (Gh)       & 3,233                               & 71.78\%          & 60.90\%                      & & 3,178                               & 70.56\%          & 59.86\%          \\
		FCN32s                & 1,746                               & 60.04\%          & 32.89\%                      & & 1,684                               & 57.91\%          & 31.72\%          \\
		FCN16s                & 3,026                               & 63.79\%          & 57.00\%                      & & 2,915                               & 61.45\%          & 54.91\%          \\
		FCN8s                 & 3,253                               & 65.27\%          & 61.27\%                      & & 3,160                               & 63.40\%          & 59.52\%          \\
		ICNet                 & 2,116                               & 68.66\%          & 39.86\%                      & & 2,054                               & 66.65\%          & 38.69\%          \\
		ENet                  & 2,999                               & 70.83\%          & 56.49\%                      & & 2,941                               & 69.46\%          & 55.40\%          \\
		U-Net                 & 3,397                               & 67.00\%          & 63.98\%                      & & 3,301                               & 65.11\%          & 62.18\%          \\
		PSPNet                & \textbf{3,408}                      & 72.50\%          & \textbf{64.19\%}             & & \textbf{3,322}                      & 70.67\%          & \textbf{62.57\%} \\ \bottomrule
	\end{tabular}
	\vspace{-4mm}
\end{table*}

\section{Experimental Results}\label{sec_experiment}

In this section, we detail the basic network implementations in terms of button segmentation and recognition, which benchmark future methods and facilitate related studies in this research field.

\subsection{Button Segmentation}\label{sec_exp_seg}

For elevator button segmentation, several popular semantic segmentation algorithms are adopted and implemented in this work. Ronneberger \textit{et al.} \cite{ronneberger2015u} design a U-shape convolutional neural network (U-Net), in which the contracting path aims to capture low-level features in shallow layers and high-level features in deeper layers, and the expansive path maps back the extracted features to the same size of the input data to reconstruct the pixel-wise segmentation mask. Long \textit{et al.} \cite{long2015fully} propose an encoder-decoder network architecture (FCN) for segmentation tasks, which replaces the fully connected (FC) layers of CNNs with convolutional layers. Based on the basic FCN, several modified versions (FCN8s, FCN16s, FCN32s) are proposed to fuse predictions of decoder layers and further improve the segmentation accuracy. Chen \textit{et al.} \cite{chen2017deeplab, chen2018encoder} publish a series of papers introducing DeepLab and its improved versions. The last proposed DeepLabv3+ is still one of the state-of-the-art (SOTA) algorithms in semantic segmentation. Zhao \textit{et al.} \cite{zhao2017pyramid} develop an effective pyramid scene parsing network (PSPNet) that is embedded with a novel pyramid pooling module for semantic segmentation. With the pyramid pooling module, PSPNet can effectively exploit the contextual information of the given input, and achieves superior performance in ImageNet scene parsing challenge 2016, PASCAL VOC 2012 benchmark, and Cityscapes benchmark. Moreover, two computationally-efficient and light-weighted neural networks, i.e., ENet \cite{paszke2016enet} and ICNet \cite{zhao2018icnet}, are adopted in this work, which may perform semantic segmentation in a real-time manner.

%These two network architectures can make a trade-off between the segmentation accuracy and processing time of a network, and perform pixel-wise semantic segmentation in real-time manner.

All algorithms mentioned above are implemented. Particularly, for DeepLab series, we adopt the latest version (DeepLabv3+) embedded with multiple feature extraction backbones, i.e., ResNet50 \cite{he2016deep}, MobileNetV2 \cite{sandler2018mobilenetv2}, MobileNetV3 \cite{howard2019searching}, and GhostNet \cite{han2020ghostnet}. The reason why we choose these backbones is due to their compact designs. We randomly split the dataset into train, validation, and test set, following a ratio of 70\%, 15\%, and 15\%. Some common augmentation strategies are adopted, i.e., random horizontal and vertical flipping, clockwise-rotation with an angle $[0^{\circ},\ 90^{\circ}]$.

During model training, the focus loss \cite{lin2017focal} is adopted, which emphasizes more on hard and misclassified button examples. The loss function is represented as follows:
\begin{equation} \label{fn_loss}
\begin{split}
l(i,j) = -g(i, j)[1-S(p(i, j))^\gamma]log[S(p(i, j))],
\end{split}
\end{equation}
where $p(i, j)$ indicates the predicted probability of pixel $(i, j)$ being categorized as the button pixel; $\gamma \geq 0$ means the focusing parameter; $S(\cdot)$ is the soft-max function; $g(i, j)\in$\{1, 0\} represents the ground-truth label. $1$ means that $(i, j)$ is the button pixel, while $0$ means the background pixel.

The results of all semantic segmentation models are summarized in Table~\ref{table:seg}. It can be seen that FCN8s, with the largest parameter size or computation complexity, achieves the highest IoU score. ENet contains almost 400 times smaller model size compared with FCN8s, but its IoU and Precision scores are quite low. PSPNet and DeeplabV3+ with ResNet50 or GhostNet backbone achieve a balance between precision and recall scores, indicating fewer efforts are required to remove false negatives and false positives of their segmentation maps. U-Net achieves a satisfactory button segmentation performance, and meanwhile, the number of trained parameters is small. Generally speaking, FCN8s and FCN16s are good solutions for button segmentation tasks concerning only the segmentation performance in the autonomous system embedded with enough computational resources, especially GPU memory. However, for systems that require real-time segmentation or have limited computation resources, U-Net is a better alternative.
%For algorithms getting higher precision normally suffer from a lower recall rate.	

In all segmentation experiments, the initial learning rate is 7e-4. Each network is trained for 50 epochs, where the model with the highest IoU in evaluation set is selected for inference. The Adam optimizer is adopted with a weight decay of 5e-4. The networks are implemented based on the PyTorch framework and trained on a workstation with Intel Core(TM) i7-10700K@3.80GHz processors and a NVIDIA GeForce RTX 2080 (8 GB) installed.

\subsection{Button Recognition}\label{sec_exp_rec}

Before conducting button recognition, the box exactly bounding each elevator button is generated based on the button segmentation masks derived in Section~\ref{sec_exp_seg}. Specifically, a powerful clustering method called density-based spatial clustering of applications with noise (DBSCAN) \cite{ester1996density} is applied on the segmentation mask first, which aims to group pixels close to each other as an independent button. The maximum and minimum coordinates of the pixels of one button are then derived, which can form one bounding box. Ideally, each bounding box contains one button. Finally, an accurate and efficient OCR framework proposed in our early work \cite{zhu2018novel} is applied to perform button character recognition, which is based on an attention-RNN (see Fig.~\ref{fig_framework}).

%Our OCR model integrates the segmentation and recognition of characters into a single neural network, which is light-weight and more robust to noises.
Following different segmentation strategies, the button recognition results on test set are shown in Table~\ref{table:recog}. According to the different designs of attention RNN, we divide the OCR model into two categories, i.e., additive and multiplicative models. Froms Table~\ref{table:recog}, we observe that the OCR models with additive attention perform slightly better than those with multiplicative attention mechanism. While testing on the whole test set, it can be seen that a highest accuracy of $64.19\%$ and $62.57\%$ is achieved in two OCR models based on PSPNet segmentation. While testing on the correctly detected buttons, Deeplabv3+ with ResNet50 backbone has better recognition results, but the number of detected buttons it leverages is smaller than that of PSPNet. U-Net achieves almost the same performance as PSPNet, and meanwhile requires much less computation resources (see Table~\ref{table:seg}), thus it can be regarded as a great candidate in elevator button recognition.

%it's still not recommended to implement PSPNet on an autonomous system for elevator manipulation. Because the processing speed and required computational resources far more than the combinations achieving a few percent less in accuracy, such as Deeplabv3+ with GhostNet and U-Net.
In all recognition experiments, the initial learning rate is set to 1e-3 and decays by 0.1x every 17 epochs, with a total of 50 training epochs. The RMSprop optimizer is adopted with a weight decay of 0.1. The networks are implemented based on the PyTorch framework and trained on a workstation with Intel Core(TM) i7-5930K@3.50GHz processors and a NVIDIA GTX TITAN X (12 GB) installed.

\section{CONCLUSIONS}\label{sec_conclusion}
In this paper, the first large-scale publicly available elevator panel dataset is released, which contains 3,718 panel images with 35,100 button labels. This dataset aims to benchmark future methods on elevator button segmentation and recognition. Along with the dataset, several measurement metrics are established for performance evaluation, and many popular network implementations for button segmentation and recognition are evaluated and released, which are believed to further push forward related studies in this research area and facilitate real-world autonomous elevator operation.

% \addtolength{\textheight}{-12cm}

\bibliographystyle{IEEEtran}
\bibliography{reference}

\end{document}